\documentclass[10pt,twocolumn,letterpaper]{article}

\usepackage{cvpr}            

\usepackage{graphicx}
\usepackage{amsmath}
\usepackage{amssymb}
\usepackage{booktabs}
\usepackage{multirow}
\usepackage{mmstyles}

\newcommand{\sggpoint}{$\text{SGG}_{\text{point}}$}

\usepackage[pagebackref,breaklinks,colorlinks]{hyperref}

\usepackage[capitalize]{cleveref}
\crefname{section}{Sec.}{Secs.}
\Crefname{section}{Section}{Sections}
\Crefname{table}{Table}{Tables}
\crefname{table}{Tab.}{Tabs.}

\def\cvprPaperID{1977} 
\def\confName{CVPR}
\def\confYear{2023}

\begin{document}

\title{VL-SAT: Visual-Linguistic Semantics Assisted Training for 3D Semantic Scene Graph Prediction in Point Cloud}

\author{
    Ziqin Wang$^1$,~~
    Bowen Cheng$^1$,~~
    Lichen Zhao$^1$,~~
    Dong Xu$^2$,~~
    Yang Tang$^{3*}$,~~
    Lu Sheng$^1$\thanks{Lu Sheng and Yang Tang are the corresponding authors.}~~ \\
    $^1$School of Software, Beihang University\\
    $^2$The University of Hong Kong, ~~ $^3$East China University of Science and Technology \\
    \small
    \texttt{\{wzqin,chengbowen052,zlc1114,lsheng\}@buaa.edu.cn} \hspace{10pt}
    \texttt{dongxu@cs.hku.hk} \hspace{10pt}
    \texttt{yangtang@ecust.edu.cn}
}
\maketitle

\begin{abstract}

The task of 3D semantic scene graph (3DSSG) prediction in the point cloud is challenging since (1) the 3D point cloud only captures geometric structures with limited semantics compared to 2D images, and (2) long-tailed relation distribution inherently hinders the learning of unbiased prediction.
Since 2D images provide rich semantics and scene graphs are in nature coped with languages, in this study, we propose \textbf{V}isual-\textbf{L}inguistic \textbf{S}emantics \textbf{A}ssisted \textbf{T}raining (\textbf{VL-SAT}) scheme that can significantly empower 3DSSG prediction models with discrimination about long-tailed and ambiguous semantic relations.
The key idea is to train a powerful multi-modal oracle model to assist the 3D model. This oracle learns reliable structural representations based on semantics from vision, language, and 3D geometry, and its benefits can be heterogeneously passed to the 3D model during the training stage.
By effectively utilizing visual-linguistic semantics in training, our VL-SAT can significantly boost common 3DSSG prediction models, such as SGFN and $\text{SGG}_\text{point}$, only with 3D inputs in the inference stage, especially when dealing with tail relation triplets.
Comprehensive evaluations and ablation studies on the 3DSSG dataset have validated the effectiveness of the proposed scheme.
%
%
Code is available at \href{https://github.com/wz7in/CVPR2023-VLSAT}{https://github.com/wz7in/CVPR2023-VLSAT}.
\end{abstract}


\section{Introduction}
\label{sec:intro}

Structurally understanding 3D geometric scenes is particularly important for tasks that require interaction with real-world environments, such as AR/VR~\cite{qi2019deep,zhao2021transformer3d,cheng2021back,shi2020points,shi2019pointrcnn,shi2020pv,zhang2020h3dnet} and navigation~\cite{cai20223djcg,he2021transrefer3d,chen2020scanrefer}.
As one vital topic in this field, predicting 3D semantic scene graph (3DSSG) in point cloud~\cite{wald2020learning} has received emerging attention in recent years.
%
%
Specifically, given the point cloud of a 3D scene that is associated with class-agnostic 3D instance masks, the 3DSSG prediction task would like to construct a directed graph whose nodes are semantic labels of the 3D instances and the edges recognize the directional semantic or geometrical relations between connected 3D instances.

%


\begin{figure}[t]
\centering
\includegraphics[width=1.0\linewidth]{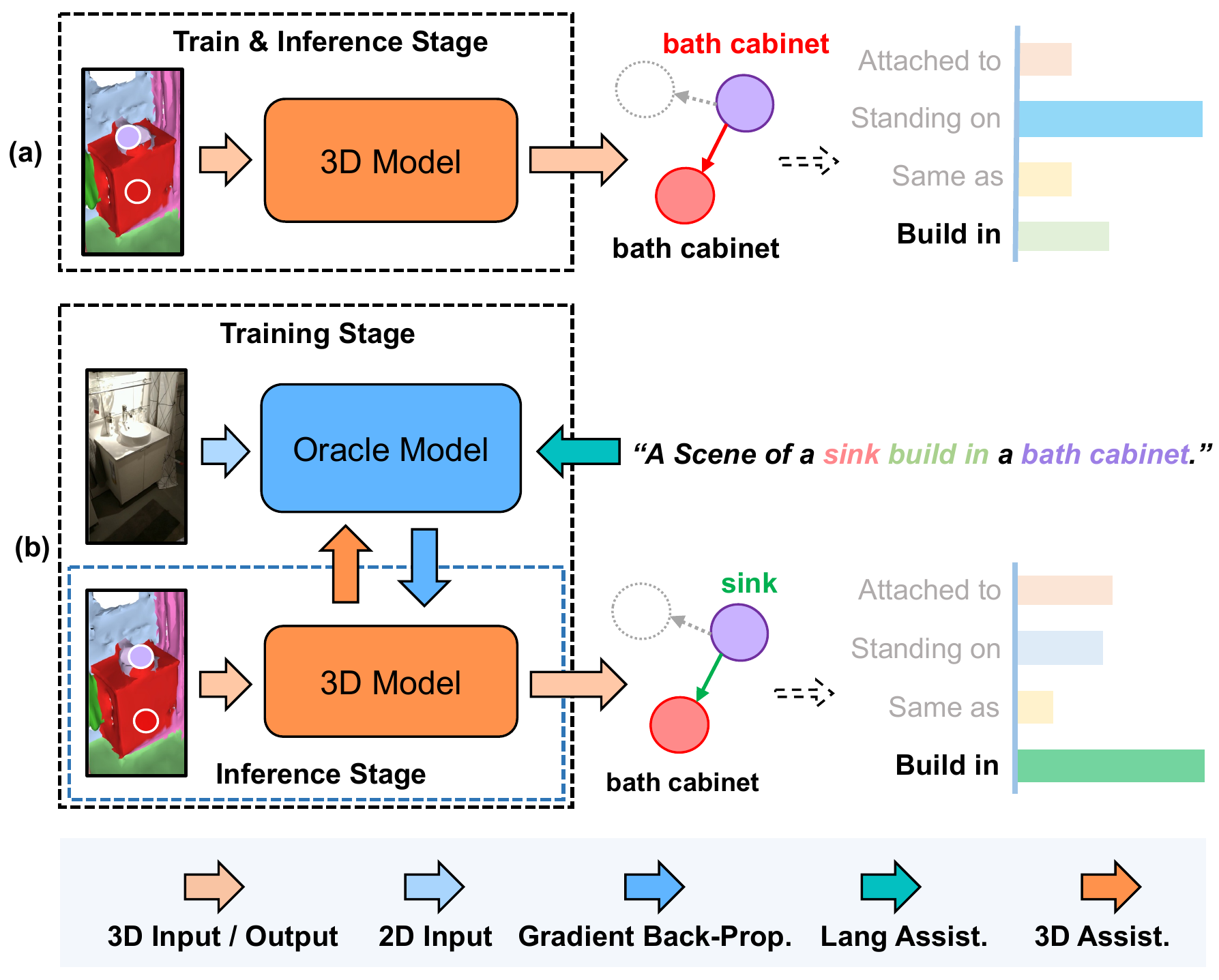}
\caption{
\textbf{Comparison between previous method and our VL-SAT.}
(a) SGPN~\cite{wald2020learning}, as the 3D model, fails to find capture predicates such as \emph{build in}.
(b) VL-SAT creates an oracle model by heterogeneously fusing 2D semantics, and language knowledge along with the geometrical features, and the 3D model receives benefits from the oracle model during training.
During inference, the enhanced 3D model can correctly detect the tail predicates.}
\label{fig:teaser}
\end{figure}

However, in addition to common difficulties faced by scene graph prediction, there are several challenges specified to the 3DSSG prediction task. (1) 3D data such as point clouds only capture the geometric structures of each instance and may superficially define the relations by relative orientations or distances.
(2) Recent 3DSSG predication datasets~\cite{wald2020learning,zhang2021exploiting} are quite small and suffer from long-tailed predicate distributions, where semantic predicates are often rarer than geometrical predicates.
For example, as shown in \cref{fig:teaser}, the pioneering work SGPN~\cite{wald2020learning} usually prefers a simple and common geometric predicate \emph{standing on} between \emph{sink} and \emph{bath cabinet}, while the ground-truth relation triplet $\langle \emph{sink}, \emph{build in}, \emph{bath cabinet}\rangle$ cares more about the semantics, and the frequency of \emph{build in} in the training dataset is quite low, as shown in \cref{fig:predicate_dis}(a).
Even though some attempts~\cite{zhang2021knowledge,wu2021scenegraphfusion,zhang2021exploiting} have been proposed thereafter, the inherent limitations of the point cloud data to some extent hinder the effectiveness of these methods.

Since 2D images provide rich and meaningful semantics, and the scene graph prediction task is in nature aligned with natural languages, we explore using visual-linguistic semantics to assist the training, as another pathway to essentially enhance the capability of common 3DSSG prediction models with the aforementioned challenges.

How to assist 3D structural understanding with visual-linguistic semantics remains an open problem.
Previous studies mainly focus on employing 2D semantics to enhance instance-level tasks, such as object detection~\cite{qi2018frustum,qi2020imvotenet,bai2022transfusion,sindagi2019mvx}, visual grounding and dense captioning~\cite{zhao20213dvg,cai20223djcg,chen2020scanrefer,yuan2022x}.
Most of them require visual data both in training and inference, but a few of them, such as SAT~\cite{yang2021sat} and $\cX$-Trans2Cap~\cite{yuan2022x} treat 2D semantics as auxiliary training signals and thus offer more practical inference only with 3D data.
But these methods are specified to instance-level tasks and require delicately designed networks for effective assistance, thus they are less desirable to our structural prediction problem. 
Thanks to the recent success of large-scale cross-modal pretraining like CLIP~\cite{radford2021learning}, 2D semantics in images can be well aligned with linguistic semantics in natural languages, and the visual-linguistic semantics have been applied for alleviating long-tailed issue in tasks related to 2D scene graphs~\cite{zareian2020bridging, abdelkarim2021exploring, tang2019learning, tang2020unbiased} and human-object interaction~\cite{liao2022gen}.
But how to adapt similar assistance of visual-linguistic semantics to the 3D scenario remains unclear.

In this study, we propose the Visual-Linguistic Semantics Assisted Training (VL-SAT) scheme to empower the point cloud-based 3DSSG prediction model (termed as the 3D model) with sufficient discrimination about long-tailed and ambiguous semantic relation triplets.
In this scheme, we simultaneously train a powerful multi-modal prediction model as the oracle (termed as oracle model) that is heterogeneously aligned with the 3D model, which captures reliable structural semantics by extra data from vision, extra training signals from language, as well as the geometrical features from the 3D model. These introduced visual-linguistic semantics have been aligned by CLIP.
Consequently, the benefits of the oracle model, especially the multi-modal structural semantics, can be efficiently embedded into the 3D model through the back-propagated gradient flows.
In the inference stage, the 3D model can perform superior 3DSSG prediction performance with only 3D inputs. For example, in \cref{fig:teaser}(b), the predicate \emph{build in} can be reliably detected.
To our best knowledge, VL-SAT is the first visual-linguistic knowledge transfer work that is applied to 3DSSG prediction in the point cloud.
Moreover, VL-SAT can successfully enhance  SGFN~\cite{wu2021scenegraphfusion} and $\text{SGG}_\text{point}$~\cite{zhang2021exploiting}, validating that this scheme is generalizable to common 3DSSG prediction models.

We benchmark VL-SAT on the 3DSSG dataset~\cite{wald2020learning}. Quantitative and qualitative evaluations, as well as comprehensive ablation studies, validate that the proposed training scheme leads to significant performance gains, especially for tail relations, as discussed in \cref{sec:experiments_and_discussions}.

\section{Related Work}
\label{sec:related_work}

\noindent\textbf{Scene Graph Prediction in Point Cloud.}
Image-based semantic scene graph prediction has been extensively studied~\cite{zellers2018neural,yang2018graph,xu2017scene,tang2020unbiased,suhail2021energy,woo2018linknet,chen2019knowledge,tang2019learning}in recent years, but only a few works try to predict 3D semantic scene graph in the point cloud.
%
%
%
Armeni~\etal~\cite{armeni20193d} presented the first 3D scene graph dataset, which maps 3D buildings into hierarchical structures.
Wald~\etal~\cite{wald2020learning} constructed a point cloud-based semantic scene graph dataset, namely 3DSSG, with a GNN-based baseline model named SGPN. 
The follow-up work SGFN~\cite{wu2021scenegraphfusion} predicted 3DSSG incrementally from RGB-D sequences.
In recent years, a few methods were proposed to improve the GNN-based baseline.
%
%
\sggpoint~\cite{zhang2021exploiting} used an edge-oriented graph convolution network to exploit multi-dimensional edge features for relation modeling.
Zhang~\etal~\cite{zhang2021knowledge} proposed a graph auto-encoder network to automatically learn a group of embeddings for each class in advance, and then perform the 3DSSG prediction to recognize credible relation triplets from pre-learned knowledge.
%

\vspace{+1mm}
\noindent\textbf{3D Scene Understanding with 2D Semantics.}
%
%
%
A list of methods have employed 2D semantics to help 3D instance-level tasks, such as 3D object detection, segmentation, visual grounding and dense captioning~\cite{qi2018frustum,qi2020imvotenet,bai2022transfusion,sindagi2019mvx,vora2020pointpainting,yin2021multimodal,zhi2021place}. They can be coarsely divided into two categories, \ie, concatenating image features with each 3D point~\cite{chen2020scanrefer, zhao20213dvg, cai20223djcg, bai2022transfusion,vora2020pointpainting,yin2021multimodal,qi2020imvotenet}, and projecting object detection results into 3D space~\cite{pang2020clocs, yang2021sat, yuan2022x, qi2018frustum, xu2018pointfusion, lahoud20172d}.
Most methods require 2D semantics both in the training and inference stages.
Recently, SAT~\cite{yang2021sat} and $\cX$-Trans2Cap~\cite{yuan2022x} explore using 2D semantics only in training to assist 3D visual grounding and dense captioning. Both of them can learn enhanced models that only use 3D inputs in inference.
But these methods are restrained to instance-level tasks and the networks have to be carefully designed.
%
%
We follow similar ideas as~\cite{yang2021sat,yuan2022x} and use 2D semantics only in training, but we would like to enhance the 3DSSG prediction that requires structural understanding rather than instance-level perception.

\vspace{+1mm}
\noindent\textbf{Knowledge-inserted Methods in Scene Graph Prediction.}
%
%
Zellers~\etal~\cite{zellers2018neural} and Chen~\etal~\cite{chen2019knowledge} indicated that the statistical co-occurrences between object pairs and relationships are useful for relation prediction.
%
%
Besides, \cite{sharifzadeh2021classification, zhang2021knowledge} generated class-level prototypical representations from all previous perceptual outputs as the prior knowledge.
These methods explicitly encoded the data priors into the model.
%
%
%
\cite{li2017vip, lu2016visual, liao2019natural, abdelkarim2021exploring} attempted to combine language priors with scene graph prediction.
%
%
Zareian~\etal~\cite{zareian2020bridging} proposed a Graph Bridging Network to propagate messages between scene graphs and knowledge graphs.
%
%
Our VL-SAT scheme uses CLIP to encode the linguistic semantics, which is thus better aligned with 2D semantics, and even the required 3D structural semantics during the training stage.
%

\section{Method}
\label{sec:method}

\begin{figure*}[t]
\centering
\includegraphics[width=1.0\linewidth]{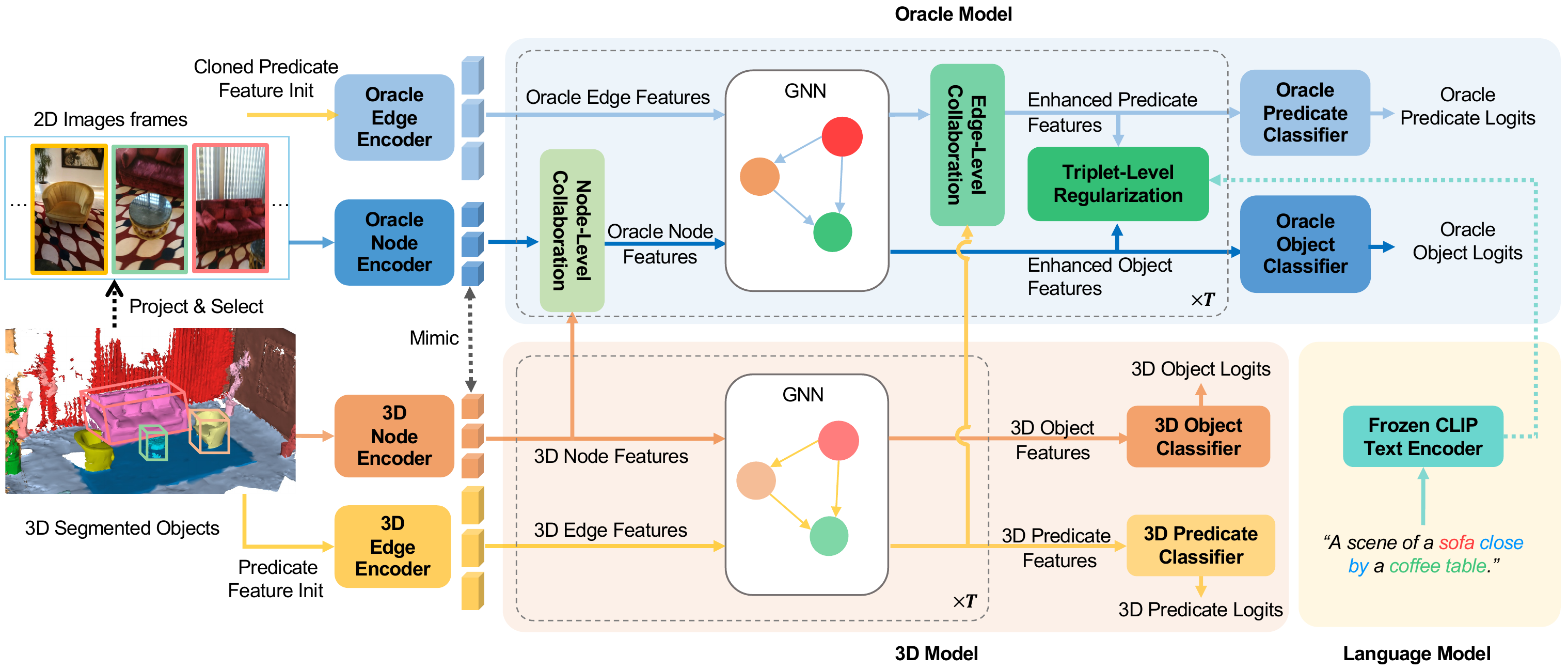}
\caption{
\textbf{The proposed Visual-Linguistic Semantics Assisted Training (VL-SAT) for 3D scene graph prediction.}
In training, VL-SAT takes 2D and language semantics as extra inputs and helps 3D scene graph prediction with node and edge-level collaboration and triplet-level regularization.
In inference, VL-SAT only takes the 3D point cloud to predict reliable 3D scene graphs.
%
%
%
}
\label{fig:network}
\end{figure*}

We first overview the formulation of 3D semantic scene graph (3DSSG) prediction in point cloud (\cref{sub:problem_formulation}) and then elaborate on a GNN-based network that we experiment on as our 3D prediction model (\cref{sub:baseline}).
Since then we highlight how our Visual-Linguistic Semantics Assisted Training (VL-SAT) scheme comprehensively transfers the benefits of an oracle multi-modal prediction model to the 3D prediction model in discriminating challenging relations (\cref{sub:heterogeneous_collaborative_learning}).
Finally, we depict the training objective in \cref{sub:training_and_inference}.

\subsection{Problem Formulation}
\label{sub:problem_formulation}

Suppose we have a point cloud $\mP \in \Rbb^{N\times3}$ with $N$ 3D points, and a set of \emph{class-agnostic} instance masks $\cM = \{\mM_1,...,\mM_K\}$ that associate the point cloud $\mP$ with $K$ semantic instances, as indicated by SGPN~\cite{wald2020learning}, we aim at predicting a 3D semantic scene graph as a directed graph $\cG = \{ \cO, \cR \}$.
The set of objects $\cO = \{ o_i \}_{i=1}^K$ are all \emph{named} object instances that are specified by instance masks $\cM$. Each edge $r_{ij}$ in $\cR$ depicts the \emph{predicate} in a relation triplet $\langle \emph{subject}, \emph{predicate}, \emph{object} \rangle$, where the head node $o_i$ of this edge is the \emph{subject} and the tail node $o_j$ is the \emph{object}.
To be specific, $o_i$ indicates an object label from $N_\text{obj}$ semantic classes.
$r_{ij}$ is a predicate label from $N_\text{rel}$ predicate classes.
%
%

\subsection{3D Prediction Model}
\label{sub:baseline}

As depicted in \cref{fig:network}, our employed 3D prediction model shares a similar network structure as those GNN-based scene graph prediction methods, such as SGFN~\cite{wu2021scenegraphfusion} and $\text{SGG}_\text{point}$~\cite{zhang2021exploiting}, which mainly consists of node encoder, edge encoder, and scene graph reasoning modules.

\vspace{+1mm}
\noindent\textbf{Node Encoder.}
Based on one class-agnostic instance mask $\mM_i$ along with the input point cloud $\mP$, we can extract the set of points $\mP_i$ that correspond to one semantic instance. 
%
%
We employ a simple PointNet~\cite{qi2017pointnet} to extract instance-level features.
The node features $\vo_i^\text{3d} \in \Rbb^{D}$ before the GNN-based scene graph reasoning are thus given by these instance-level features.

\vspace{+1mm}
\noindent\textbf{Edge Encoder.}
We follow the same practice as in SGFN~\cite{wu2021scenegraphfusion} to encode the edge features for the GNN-based scene graph reasoning.
It requires calculating the differences between several attributes between the linked instances. 
For each instance, these attributes include the mean $\vmu$ and standard deviation $\vsigma$ of the 3D points, the size $\vb = (b_x, b_y, b_z)$, the volumn $v = b_xb_yb_z$, and the maximum side length $l=\max(b_x, b_y, b_z)$ of the bounding box.
Thus the edge features $\vr_{ij}^\text{3d}\in\Rbb^D$ are encoded by projecting the concatenated differences of these attributes between two instances, via multi-layer perceptron (MLP) layers, \ie,
\begin{equation}
    \vr_{ij}^\text{3d} = \mathtt{MLP}(\mathtt{cat}(\vmu_i-\vmu_j, \vsigma_i-\vsigma_j,\vb_i-\vb_j,\ln\frac{l_i}{l_j},\ln \frac{v_i}{v_j})),
\end{equation}
where the subscript $i$ indicates the instance $\mP_i$ in the head node, and the $j$ means the instance $\mP_j$ in the tail node.

\vspace{+1mm}
\noindent\textbf{Scene Graph Reasoning.}
In our experiment, we apply a similar GNN structure as in SGFN~\cite{wu2021scenegraphfusion}, which utilizes a Feature-wise Attention (FAT) module~\cite{wu2021scenegraphfusion} to pass messages between nodes and edges, and then gets the updated node and edge features.
Each GNN module is paired with a multi-head self-attention (MHSA) module, and they are repeated for $T$ times to extract the final node and edge features $\{\tilde{\vo}_i^\text{3d}\}_{i=1,\ldots,K}$ and $\{\tilde{\vr}_{ij}^\text{3d}\}_{i\neq j, i,j=1,\ldots,K}$.
%
%
Since then, an object classifier and a predicate classifier are to predict the elements $\{ o_i, r_{ij}, o_j \}$ of each possible relation triplet from the triplet features $\{ \tilde{\vo}_i^\text{3d}, \tilde{\vr}_{ij}^\text{3d}, \tilde{\vo}_j^\text{3d} \}$.
These relation triplets finally construct the semantic scene graph $\cG = \{ \cQ, \cR \}$.

\vspace{+1mm}
Note that the 3D prediction model does not have to strictly follow the network of SGFN~\cite{wu2021scenegraphfusion}. More recent GNN-based models, such as~$\text{SGG}_\text{point}$~\cite{zhang2021exploiting} can also be applied. In \cref{subsec:ablation}, we show that the proposed VL-SAT scheme can also enhance both baselines with significant gains, validating that our method is generalizable to common 3DSSG prediction models.

\subsection{Visual-Linguistic Semantics Assisted Training}
\label{sub:heterogeneous_collaborative_learning}

In this subsection, we elaborate on how the visual-linguistic semantics assisted training (VL-SAT) scheme can empower the 3D prediction model with sufficient discrimination about long-tailed and ambiguous semantic relation triplets.
The key idea is that this discriminative power comes from auxiliarily learning a powerful multi-modal prediction model that receives structural semantics from vision, and language, as well as the 3D geometry from the 3D prediction model.
The multi-modal semantics are expected to be heterogeneously aligned with the 3D semantics at the node and edge levels, and the benefits from the oracle model can be efficiently absorbed by the 3D prediction model during the training process.
To be specific, we first introduce the applied multi-modal prediction model that has heterogeneous collaboration with the 3D prediction model, and then the auxiliary training strategies that boost the performance of the oracle model and eventually enhance the 3D prediction model.
We present the pipeline of VL-SAT in \cref{fig:network}.

\vspace{+1mm}
\noindent\textbf{Multi-modal Prediction Model as the Oracle.}
This multi-modal prediction model acts as the oracle to our 3D prediction model. It copies the 3D prediction network in~\cref{sub:baseline} and also learns to predict 3D semantic scene graphs, but its node features are represented by visual features. 
These visual features are extracted by a fixed 2D instance encoder, describing RGB image patches that are associated with each point cloud instance $\mP_i$\footnote{Please refer to the supplementary for the details about how to gather associated image patches with each point cloud instance.}. 
The edge features of the multi-modal prediction model are encoded in the same way in \cref{sub:baseline}, thus still capturing 3D spatial structures in understanding relations.

The features of this oracle model heterogeneously collaborate with those in the 3D prediction model at the node and edge levels, which are conducted before and after each GNN layer in the scene graph reasoning module.
The former is a node-level collaboration, and the latter is an edge-level collaboration.
To be specific, these collaboration operations are implemented by multi-head cross-attention (MHCA) modules~\cite{vaswani2017attention}, where the keys and values are node/edge features from the 3D model, and the queries are their counterparts from the multi-modal model.
The node-level collaboration has a distance-aware masking strategy to remove unnecessary attention between instances that are far apart without valid relations.
The mask value between two instances $\mP_i$ and $\mP_j$ are learned by
\begin{equation}
    D_{ij}^\text{node} = \mathtt{MLP}(\mathtt{cat}(\vmu_i-\vmu_j, \| \vmu_i-\vmu_j\|_2)),
\end{equation}
with respect to the mean coordinates $\vmu_i$ and $\vmu_j$ of the point cloud instances $\mP_i$ and $\mP_j$.
The edge-level collaboration does not use a distance-aware masking strategy since the distance between edges is hard to define, thus it is safer to incorporate all edges into attention calculation. 

Note that the heterogeneous collaboration is \emph{unidirectional} from the 3D model to the oracle model, while the benefits of the oracle model are passed to the 3D model through the back-propagated gradient flows.
It favors that in the inference stage, predicting 3D semantic scene graphs will not need extra data from other modalities.

\vspace{+1mm}
\noindent\textbf{Auxiliary Training Strategies.}
%
%
Since the oracle multi-modal model would like to perceive scene graphs from both the visual and linguistic perspective, it is natural to enhance the oracle model using visual-linguistic knowledge captured by CLIP~\cite{gao2021clip}.
Specifically, we can generate CLIP text embedding $\ve^{\text{text}}_{ij}$ for each groundtruth relation triplet, and regularize the corresponding triplet features $\{\tilde{\vo}^{\text{oracle}}_i, \tilde{\vr}^{\text{oracle}}_{ij}, \tilde{\vo}^{\text{oracle}}_j\}$ at the end of each GNN layer of the scene graph reasoning module.
The CLIP text embeddings are offline extracted by the template ``A scene of a/an [subject][predicate] a/an [object]'' for each GT relation 
Thus, the regularization becomes to minimize the embedding distances between the text embeddings $\ve_{ij}^\text{text}$ and the fused triplet features $\vt_{ij}^\text{oracle}$, \ie,
\begin{equation}
    L_\text{tri-emb} = \sum_{i=1}^K \sum_{j=1,j\neq i}^K \rho(\vt_{ij}^\text{oracle}, \ve_{ij}^\text{text}) \cdot \Ibb_{[\ve_{ij}^\text{text}~\text{is from GT triplet}]} \label{eq:triplet_embedding_loss}
\end{equation}
where $\vt_{ij}^\text{oracle} = \mathtt{MLP}(\mathtt{cat}(\tilde{\vo}_i^\text{oracle}, \tilde{\vr}_{ij}^\text{oracle}  ,\tilde{\vo}_j^\text{oracle}))$ is the fused embedding of the concatenated features $\tilde{\vo}_i^\text{oracle}, \tilde{\vr}_{ij}^\text{oracle}$, and  $\tilde{\vo}_j^\text{oracle}$.
$\rho(\cdot, \cdot)$ is a distance metric, we can apply $\ell_1$ norm or negative cosine distance.
$\Ibb_{[\cdot]}$ is an indicator function that equals to $1$ when the argument is true, and $0$ otherwise.
Thus \cref{eq:triplet_embedding_loss} only regularizes the node and edge features whose triplets have ground-truth relations. 

In addition, before being put into the scene graph reasoning modules, the 3D node features $\vo_i^\text{3d}$ from the 3D model and the 2D node features $\vo_i^\text{2d}$ from the oracle model can be aligned. 
We apply a same distance measurement as \cref{eq:triplet_embedding_loss},
\begin{equation}
    L_\text{node-init} = \sum_{i=1}^K \rho(\vo_i^\text{3d}, \vo_i^\text{2d}).
\end{equation}
To enhance the representation ability of the initialized 2D node features, the 2D instance encoder is a fixed CLIP-pretrained vision encoder.
Moreover, to enhance the object classifiers of both models, we use the CLIP object embeddings to initialize the weights of the object classifiers both at the 3D prediction model and the oracle multi-modal prediction model, as in~\cite{radford2021learning,liao2022gen}.

\subsection{The Training Objective}
\label{sub:training_and_inference}
%
%
The training objective of the entire network is defined as:
\begin{multline}
L = \lambda_{\text{obj}}(L_{\text{obj}}^{\text{3d}}+L_{\text{obj}}^{\text{oracle}}) + \lambda_{\text{pred}}(L_{\text{pred}}^{\text{3d}}+L_{\text{pred}}^{\text{oracle}}) + \\ \lambda_{\text{aux}}(L_{\text{tri-emb}}+L_{\text{node-init}})
\label{eq:loss_total}
\end{multline}
$L_{\text{obj}}$ indicates object classification loss and is implemented with cross-entropy loss.
$L_{\text{obj}}^{\text{3d/oracle}}$ is applied on 3D/oracle object classifier.
$L_{\text{pred}}$ indicates predicate classification loss and is formulated as per-class binary cross-entropy loss as in ~\cite{wald2020learning}.
$L_{\text{pred}}^{\text{3d/oracle}}$ is applied on 3D/oracle predicate classifier.
$\lambda_{\text{node}}$, $\lambda_{\text{edge}}$, $\lambda_{\text{aux}}$ are hyper-parameters to balance each loss in the same scale.

%

\begin{table*}
\centering
\caption{
Quantitative Results of 3D semantic scene graph prediction on the 3DSSG validation set~\cite{wald2020learning}. 
Evaluations are conducted in terms of object, predicate, and triplet. 
The results of SGPN, \sggpoint, and SGFN are based on our reproduced model with point cloud-only inputs, since they don't compute the mA@$k$ metric in their papers.
}
\resizebox{\linewidth}{!}{
\begin{tabular}{l|c|c|c|c|c|c|c|c|c|c|c|c|c}
\hline
\multirow{2}{*}{Model} & \multicolumn{3}{c|}{Object} & \multicolumn{6}{c|}{Predicate} & \multicolumn{4}{c}{Triplet} \\
 \cline{2-14}
& A@$1$ & A@$5$ & A@$10$ & A@$1$ & A@$3$ & A@$5$ & mA@$1$ & mA@$3$ & mA@$5$ & A@$50$ & A@$100$ & mA@$50$ & mA@$100$ \\
 \hline
 SGPN~\cite{wald2020learning} & 48.28 & 72.94 & 82.74 & \textbf{91.32} & 98.09 & 99.15 & 32.01 & 55.22 & 69.44 & 87.55 & 90.66 & 41.52 & 51.92 \\
\sggpoint~\cite{zhang2021exploiting} & 51.42 & 74.56 & 84.15 & 92.4 & 97.78 & 98.92 & 27.95 & 49.98 & 63.15 & 87.89 & 90.16 & 45.02 & 56.03 \\
 SGFN~\cite{wu2021scenegraphfusion} & 53.67 & 77.18 & 85.14 & 90.19 & 98.17 & 99.33 & 41.89 & 70.82 & 81.44 & 89.02 & 91.71 & 58.37 & 67.61 \\
 \hline
 non-VL-SAT & 54.79 & 77.62 & 85.84 & 89.59 & 97.63 & 99.08 & 41.99 & 70.88 & 81.67 & 88.96 & 91.37 & 59.58 & 67.75 \\
 VL-SAT (ours) & \textbf{55.66} & \textbf{78.66} & \textbf{85.91} & 89.81 & \textbf{98.45} & \textbf{99.53} & \textbf{54.03} & \textbf{77.67} & \textbf{87.65} & \textbf{90.35} & \textbf{92.89} & \textbf{65.09}  & \textbf{73.59} \\
 \hline
 VL-SAT (oracle) & 66.39 & 86.53 & 91.46 & 90.66 & 98.37 & 99.40 & 55.66 & 76.28 & 86.45 & 92.67 & 95.02 & 74.10 & 81.38 \\
 \hline
\end{tabular}}
\label{tab:exp-results}
\end{table*}

\section{Experiments and Discussions}
\label{sec:experiments_and_discussions}

\subsection{Setups and Implementation Details}
\label{subsec:exp-setup}

\noindent\textbf{Datasets.} 
We conduct experiments on 3DSSG~\cite{wald2020learning}. It is a 3D semantic scene graph dataset drawn from the 3RScan dataset~\cite{wald2019rio}, with rich annotations about instance segmentation masks and relation triplets.
It has $1553$ 3D reconstructed indoor scenes, $160$ classes of objects, and $26$ types of predicates.
%
%
%
In the experiments, we use the same data preparation and training/validation split as in 3DSSG~\cite{wald2020learning}.

\vspace{+1mm}
\noindent\textbf{Metrics and Tasks.} 
We follow the experiment settings in 3DSSG~\cite{wald2020learning} . 
In both training and testing stages, 3D scenes are placed in the same 3D coordinate. The view-dependent spatial relation predicates are not ambiguous.
%
To evaluate the prediction of the object and predicate, we use the top-k accuracy (A@$k$) metric.
As for the triplets, we first multiply the subject, predicate, and object scores to get triplet scores, and then compute the top-k accuracy (A@$k$) as the evaluation metric.
The triplet is considered correct only if the subject, predicate, and object are all correct\footnote{However, the metric top-k accuracy is written as the top-k recall or R@k in 3DSSG~\cite{wald2020learning} and SGFN~\cite{wu2021scenegraphfusion}.}.
%
To fairly evaluate the performance of long-tailed predicate distribution, we also compute the average top-k accuracy of the predicate across all predicate classes, denoted as the mean top-k accuracy (mA@$k$).
%

%
We also conduct two 2D scene graph tasks proposed in \cite{xu2017scene} in the 3D scenario, as what Zhang~\etal~\cite{zhang2021knowledge} did, \ie,
(1) Scene Graph Classification (SGCls) that \textcolor{black}{evaluates} the triplet together.
(2) Predicate Classification (PredCls) that only \textcolor{black}{evaluates} the predicate \textcolor{black}{with the ground-truth labels of object entities.}
%
Following Zhang~\etal~\cite{zhang2021knowledge}, we compute the recall at the top-k (R@$k$) triplets.
The triplet is considered correct when the subject, predicate, and object are all valid.
%
%
Additionally, we also adopt mean recall (mR@$k$) to evaluate the performance on the unevenly sampled relations using a similar strategy as mA@$k$.

\vspace{+1mm}
\noindent\textbf{Implementation Details.} 
Our network is end-to-end optimized using AdamW optimizer~\cite{DBLP:journals/corr/KingmaB14,DBLP:journals/corr/abs-1711-05101} with the batch size as $8$.
We train the network for $100$ epochs, and the base learning rate is set as $0.001$ with a cosine annealing learning rate decay strategy~\cite{loshchilov2016sgdr}.
$N_\text{obj}=160$ and $N_\text{rel}=26$ in our experiments.
GNN modules are repeated for $T=2$ times in both 3D and oracle multi-modal models.
$\lambda_{\text{obj}}=\lambda_{\text{aux}}=0.1$, $\lambda_{\text{pred}}=1$ in \cref{eq:loss_total}.
All experiments are carried out on the PyTorch platform equipped with one NVIDIA GeForce RTX 2080 Ti GPU card, and each experiment takes about $48$ hours until model convergence.
Note that 2D inputs are only used during the training stage.
During the inference stage, we follow the same strategy in  ~\cite{xu2017scene}, which selects the top@$1$ class of both object and predicate while giving an object instance index tuple. 
%
Please refer to the supplementary for the details of the network structures.

\subsection{Comparison with the State-of-the-art Methods}
\label{subsec:exp-results}
We compare our method with a list of reference methods, \ie SGPN~\cite{wald2020learning}, \sggpoint~\cite{zhang2021exploiting}, SGFN~\cite{wu2021scenegraphfusion}, Co-Occurrence~\cite{zhang2021knowledge}, KERN~\cite{chen2019knowledge}, Schemata~\cite{sharifzadeh2021classification}, Zhang~\etal~\cite{zhang2021knowledge}. 
%
%
In addition, to gain a deeper understanding of our approach, we also report the performances of the oracle multi-modal prediction model (termed as VL-SAT (oracle)), as well as the baseline performance of the 3D prediction model that is trained purely by 3D data (term as non-VL-SAT). The proposed method is term as VL-SAT.


\begin{figure}[t]
\centering
\includegraphics[width=1\linewidth]{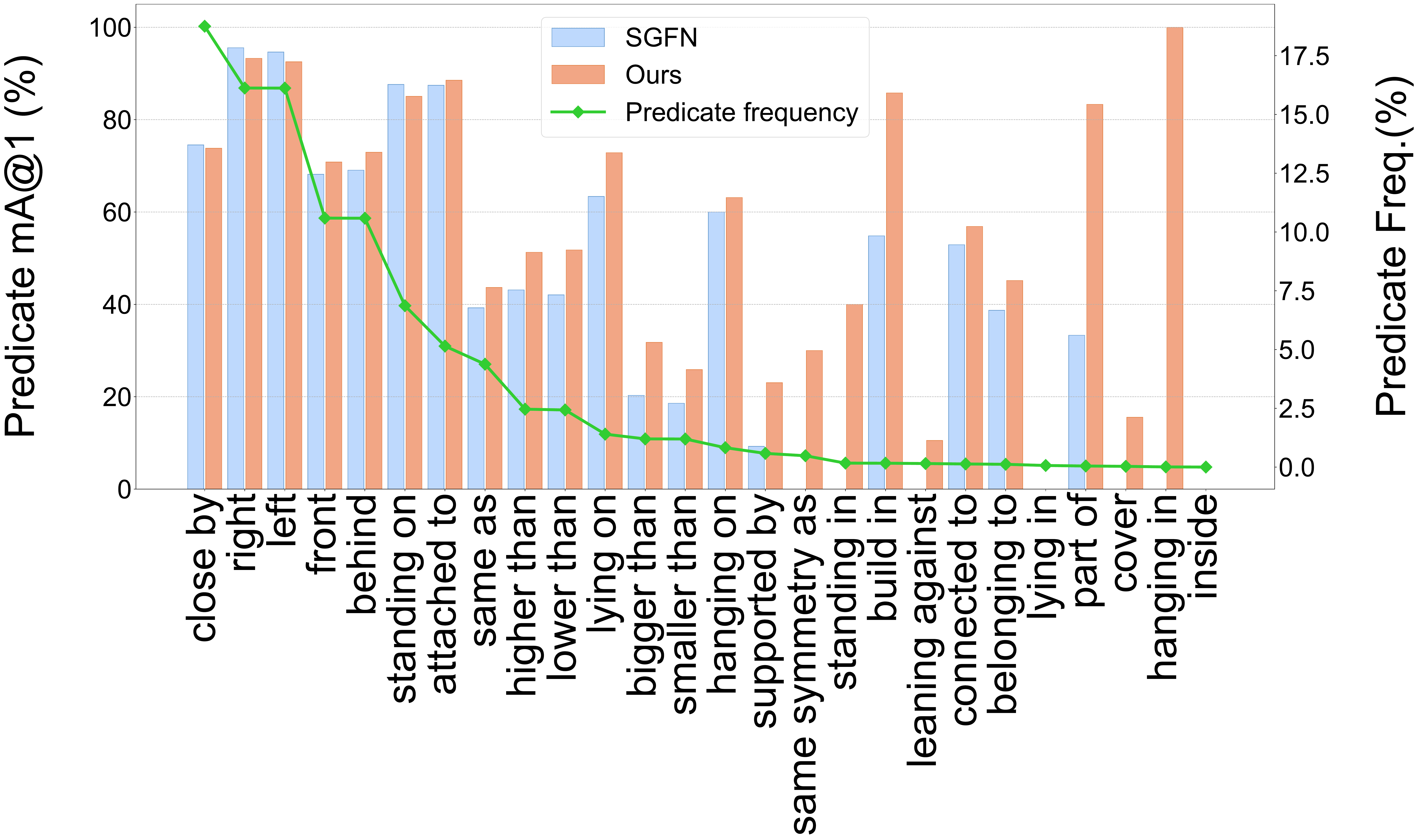}
\caption{
The line chart shows the predicate frequency in the train set of 3DSSG~\cite{wald2020learning}.
The bar chart shows the results on mA@$1$ of the predicate prediction of SGPN~\cite{wald2020learning} and our VL-SAT.
}
\label{fig:predicate_dis}
\end{figure}

\vspace{+1mm}
\noindent\textbf{Quantitative Results.}
The comparison results are summarized in \cref{tab:exp-results}.
The baseline ``non-VL-SAT'' has a similar performance as SGFN.
\textcolor{black}{The only difference between them is that ``non-VL-SAT'' adds a multi-head self-attention (MHSA) module~\cite{vaswani2017attention} before each GNN module in SGFN.}
Thanks to the delicate visual-linguistic assisted training scheme, our ``VL-SAT'' tremendously improves the baseline, according to the evaluation with respect to the predicate, and the triplet.
Moreover, according to the less biased mA@k metrics with respect to long-tailed distribution, when evaluating the predicate, the proposed ``VL-SAT'' outperforms the baseline ``non-VL-SAT'' with around $12.0\%$, $6.8\%$ and $6.0\%$ gains at mA@$1$, mA@$3$, and mA@$5$ respectively.
Our method reaches new state-of-the-art results on triplet prediction, with $6.8\%$ gain on mA@50 and $5.9\%$ gain on mA@100 over SGFN~\cite{wu2021scenegraphfusion}.
Note that the results of object classification just have a marginal improvement, which means that a simple PointNet-based 3D encoder may not be able to convey similar instance-level representative power as the 2D vision encoder.
%


As illustrated in \cref{tab:exp-buaa1} and \cref{tab:exp-buaa2}, we also compare our ``VL-SAT'' with the reference methods, with respect to two tasks named SGCls and PredCls, according to the settings introduced by Zhang~\etal~\cite{zhang2021knowledge}.
Our method outperforms Zhang~\etal~\cite{zhang2021knowledge} by a large margin.
For example, with graph constraint~\cite{xu2017scene} (as a more rigorous testing scenario~\cite{zareian2020bridging}), ``VL-SAT'' has $2.5\%$ gain on R@$20$ in SGCls, $8.5\%$ gain on R@$20$ in PredCls.
Moreover, with respect to the less biased metrics in \cref{tab:exp-buaa2}, ``VL-SAT'' even achieves $6.6\%$ gains on mR@$20$ in SGCls than Zhang~\etal~\cite{zhang2021knowledge}.

\begin{table}
\caption{
Quantitative results of the compared methods with respect to the SGCls and PredCls tasks, with and without graph constraint. The evaluation metric is top-$k$ recall.
}
\label{tab:exp-buaa1}
\centering
\resizebox{0.9\linewidth}{!}{
\begin{tabular}{c|c|c}
\hline
\multicolumn{1}{c|}{}& SGCls & PredCls \\
\cline{2-3}
Model & R@$20$/$50$/$100$ & R@$20$/$50$/$100$ \\
\midrule
\multicolumn{3}{c}{with Graph Constraints} \\
\hline
Co-Occurrence~\cite{zhang2021knowledge} & 14.8/19.7/19.9 & 34.7/47.4/47.9 \\
KERN~\cite{chen2019knowledge} & 20.3/22.4/22.7 & 46.8/55.7/56.5 \\
SGPN~\cite{wald2020learning} & 27.0/28.8/29.0 & 51.9/58.0/58.5 \\
Schemata~\cite{sharifzadeh2021classification} & 27.4/29.2/29.4 & 48.7/58.2/59.1 \\
Zhang~\etal~\cite{zhang2021knowledge} & 28.5/30.0/30.1 & 59.3/65.0/65.3 \\
SGFN~\cite{wu2021scenegraphfusion} & 29.5/31.2/31.2 & 65.9/78.8/79.6 \\
VL-SAT (ours) & \textbf{32.0}/\textbf{33.5}/\textbf{33.7} & \textbf{67.8}/\textbf{79.9}/\textbf{80.8} \\
\midrule
\multicolumn{3}{c}{without Graph Constraints} \\
\hline
Co-Occurrence~\cite{zhang2021knowledge} & 14.1/20.2/25.8 & 35.1/55.6/70.6 \\
KERN~\cite{chen2019knowledge} & 20.8/24.7/27.6 & 48.3/64.8/77.2 \\
SGPN~\cite{wald2020learning} & 28.2/32.6/35.3 & 54.5/70.1/82.4 \\
Schemata~\cite{sharifzadeh2021classification} & 28.8/33.5/36.3 & 49.6/67.1/80.2 \\
Zhang~\etal~\cite{zhang2021knowledge} & 29.8/34.3/37.0 & 62.2/78.4/88.3 \\
SGFN~\cite{wu2021scenegraphfusion} & 31.9/39.3/45.0 & 68.9/82.8/91.2 \\
VL-SAT (ours) & \textbf{33.8}/\textbf{41.3}/\textbf{47.0} & \textbf{70.5}/\textbf{85.0}/\textbf{92.5} \\
\hline
\end{tabular}}
\end{table}

\begin{table}
\centering
\caption{
Quantitative results of the compared methods with respect to the SGCls and PredCls tasks, with graph constraint. The evaluation metric is top-$k$ mean recall.
}
\label{tab:exp-buaa2}
\resizebox{0.9\linewidth}{!}{
\begin{tabular}{c|c|c}
\hline
&\multicolumn{1}{|c|}{SGCls} & \multicolumn{1}{|c}{PredCls} \\
\cline{2-3}
Model & mR@$20$/$50$/$100$ & mR@$20$/$50$/$100$ \\
\hline
Co-Occurrence~\cite{zhang2021knowledge} & 8.8/12.7/12.9 & 33.8/47.4/47.9 \\
KERN~\cite{chen2019knowledge} & 9.5/11.5/11.9 & 18.8/25.6/26.5 \\
SGPN~\cite{wald2020learning} & 19.7/22.6/23.1 & 32.1/38.4/38.9 \\
Schemata~\cite{sharifzadeh2021classification} & 23.8/27.0/27.2 & 35.2/42.6/43.3 \\
Zhang~\etal~\cite{zhang2021knowledge} & 24.4/28.6/28.8 & 56.6/63.5/63.8 \\
SGFN~\cite{wu2021scenegraphfusion} & 20.5/23.1/23.1 & 46.1/54.8/55.1 \\
VL-SAT(ours) & \textbf{31.0}/\textbf{32.6}/\textbf{32.7} & \textbf{57.8}/\textbf{64.2}/\textbf{64.3} \\
\hline 
\end{tabular}}
\end{table}

\begin{figure*}
\centering
\includegraphics[width=0.99\linewidth]{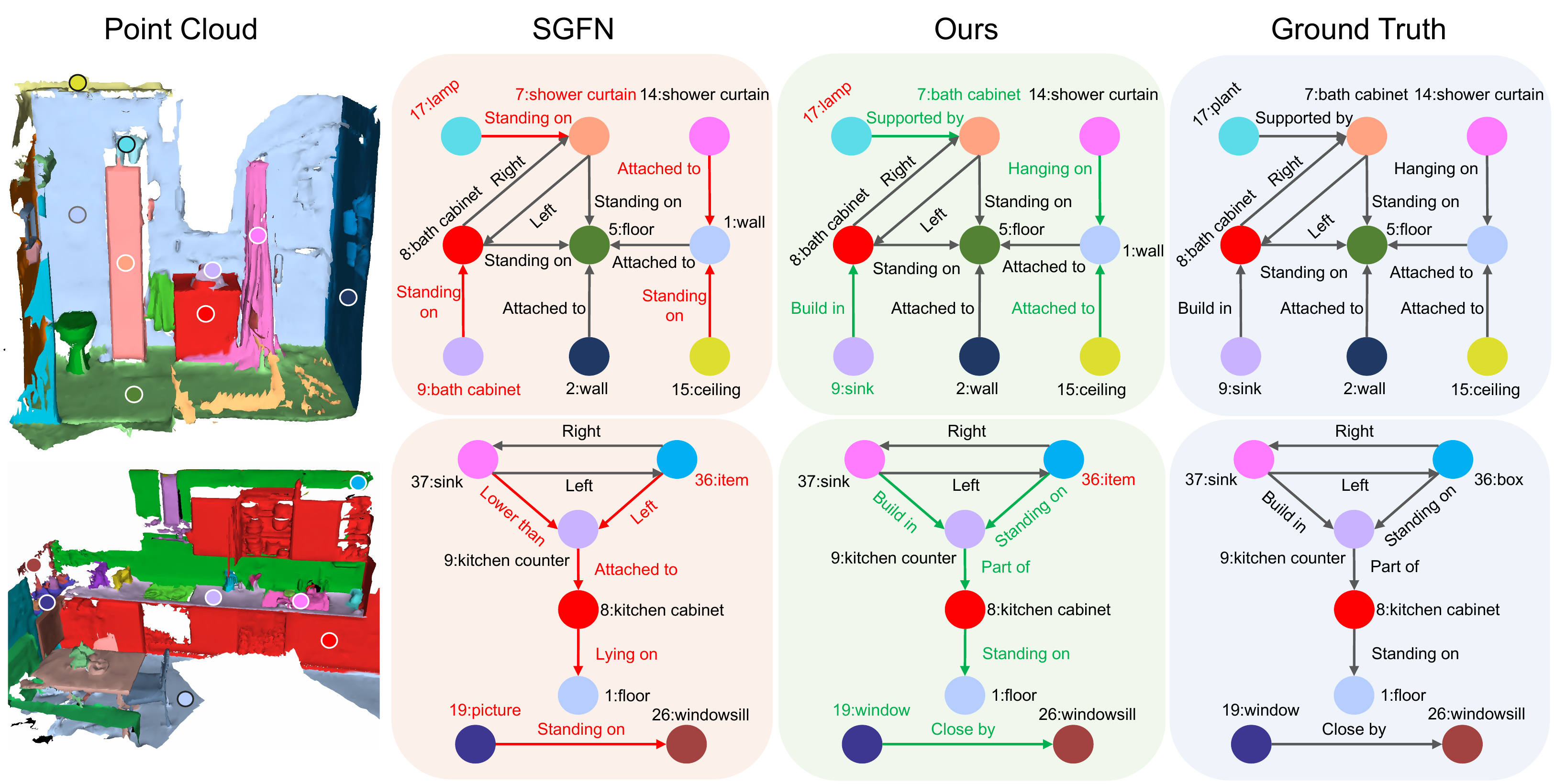}
\caption{
Qualitative results from SGFN~\cite{wu2021scenegraphfusion} and our method on the 3DSSG~\cite{wald2020learning} dataset.
\textit{Red edge}: miss-classified edges from SGFN, \textit{green edge}: edges corrected by our method, \textit{red node}: miss-classified node.
}
\label{fig:qualitative}
\end{figure*}

\begin{figure*}
\centering
\includegraphics[width=0.95\linewidth]{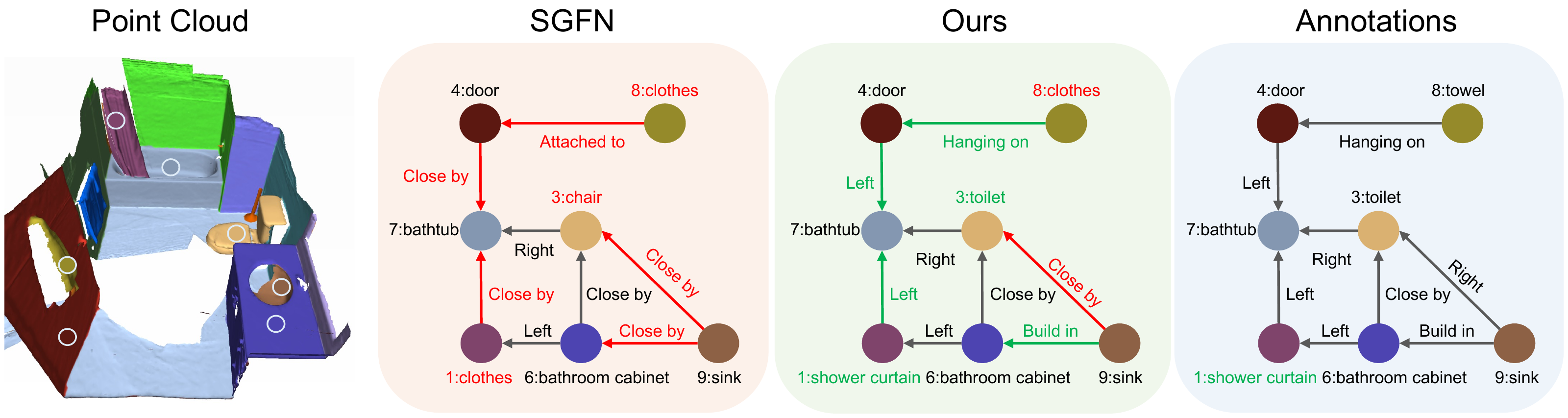}
\caption{
Qualitative results from SGFN~\cite{wu2021scenegraphfusion} and our method on ScanNet~\cite{dai2017scannet} dataset. 
Note that there are no annotations of scene graphs on ScanNet~\cite{dai2017scannet}, so we utilize the descriptions from ScanRefer~\cite{chen2020scanrefer} to parse the relationships between different objects manually.
\textit{Red edge}: miss-classified edges from SGFN, \textit{green edge}: edges corrected by our method, \textit{red node}: miss-classified node.
}
\label{fig:qualitative_scannet}
\end{figure*}

\begin{table*}
\centering
\caption{
Based on the distribution of the predicates in the train set of the 3DSSG dataset~\cite{wald2020learning}, we split the $26$ predicate classes into the head, body, and tail classes, and then compute mA@3 and mA@5 metrics on each split.
%
%
Moreover, we test several methods on unseen and seen triplets in the validation set to evaluate the generalization ability of these methods.
}
\label{tab:long-tail}
\resizebox{0.8\linewidth}{!}{
        \begin{tabular}{l|c|c|c|c|c|c|c|c|c|c}
\hline
\multirow{3}{*}{Model}& \multicolumn{6}{|c|}{Predicate} & \multicolumn{4}{|c}{Triplet} \\
\cline{2-11}
&\multicolumn{2}{c|}{Head} & \multicolumn{2}{|c|}{Body} & \multicolumn{2}{|c}{Tail} & \multicolumn{2}{|c|}{Unseen} & \multicolumn{2}{|c}{Seen} \\
 \cline{2-11}
& mA@$3$ & mA@$5$ &mA@$3$ & mA@$5$ & mA@$3$ & mA@$5$ & A@50 & A@100 & A@50 & A@100 \\
 \hline
 SGPN~\cite{wald2020learning} & \textbf{96.66} & 99.17 & 66.19 & 85.73 & 10.18 & 28.41 & 15.78 & 29.60 & 66.60 & 77.03  \\
 SGFN~\cite{wu2021scenegraphfusion} & 95.08 & \textbf{99.38} & 70.02 & 87.81 & 38.67 & 58.21 & 22.59 & 35.68 & 71.44 & 80.11 \\
 \hline
 non-VL-SAT & 95.32 & 99.01 & 71.88 & 88.64 & 40.01 & 58.33 & 21.99 & 35.44 & 71.52 & 80.34  \\
 VL-SAT (ours) & 96.31 & 99.21 & \textbf{80.03} & \textbf{93.64}  & \textbf{52.38} & \textbf{66.13} & \textbf{31.28} & \textbf{47.26} & \textbf{75.09} & \textbf{82.25} \\
 \hline
\end{tabular}}
\end{table*}

\vspace{+1mm}
\noindent\textbf{Qualitative Results.}
We provide some qualitative results between SGFN and our ``VL-SAT'' in \cref{fig:qualitative}.
These results demonstrate that our method can predict more reliable scene graphs with more accurate edges and nodes.
For example, our method successfully distinguishes some similar predicate, like \textit{standing on} versus \textit{supported by} and further disambiguates related instances, such as \textit{shower curtain} versus \textit{bath cabinet}.
The results conducted on ScanNet~\cite{dai2017scannet} in \cref{fig:qualitative_scannet} validate that VL-SAT is generalizable to more datasets.

\subsection{More Evaluations about Predicate and Triplet}
\label{subsec:exp-long-tail}

\noindent\textbf{Tail Predicates.} 
In \cref{fig:predicate_dis}, we visualize the frequency of the predicates in the train set as the line chart, which shows the long-tail distribution.
We also show the per-class predicate prediction performances of ``VL-SAT'' and SGFN~\cite{wald2020learning} in the bar chart.
Compared with SGFN, our method gets a significant improvement in the tail categories.
To further explore the improvements brought by ``VL-SAT'', we split the $26$ predicate classes into three parts: \textit{head}, \textit{body}, and \textit{tail} according to their frequencies in the train set, and calculate mA@$k$ metric.
In \cref{tab:long-tail}, we obtain $13.71\%$ improvement on mA@$3$ when predicting the predicate on tail categories.
\textcolor{black}{Compared with SGFN, our method slightly drops on some head classes but significantly increases on tail classes.
Since our VL-SAT boosts the overall performance by a large margin, such a slight performance degradation in head predicates is acceptable.}
We also provide some examples of tail predicates in \cref{fig:qualitative}.
In the top row, our method correctly predicts tail predicates like \textit{supported by}.
In the bottom row, our method corrects the relation between the kitchen counter and the kitchen cabinet from \textit{attached to} to \textit{part of}.



\vspace{+1mm}
\noindent\textbf{Unseen Triplets.} 
We consider relation triplets that do not appear in the train set as unseen triplets. 
In \cref{tab:long-tail}, our method gains about $8.69\%$ on A@$50$ on unseen triplets compared with SGFN~\cite{wald2020learning}.
The results validate that thanks to the VL-SAT scheme, our model can convey more robust feature representations based on the 3D point cloud, which leads to a better generalization ability on unseen triplets.

\begin{table}
\centering
\caption{
Results of our method when different modules are ablated.
CI means CLIP-initialized object classifier.
NC means node-level collaboration.
EC means edge-level collaboration.
TR means triplet-level CLIP-based regularization.
}
\label{tab:ablation_exp}
\resizebox{\linewidth}{!}{
\begin{tabular}{c|c|c|c|c|c|c|c|c|c}
\hline
\multirow{2}{*}{CI} & \multirow{2}{*}{NC} & \multirow{2}{*}{EC} & \multirow{2}{*}{TR} & \multicolumn{2}{|c|}{Object} & \multicolumn{2}{|c|}{Predicate} & \multicolumn{2}{|c}{Triplet} \\
\cline{5-10}
& & & & A@$5$ & A@$10$ & mA@$3$ & mA@$5$ & mA@$50$ & mA@$100$ \\
\hline
& & & & 77.62 & 85.84 & 70.88 & 81.67 & 59.58 & 67.75 \\
\checkmark & & & & 79.03 & 86.81 & 72.50 & 83.59 & 60.65 & 69.71 \\
\checkmark & \checkmark & & & \textbf{79.28} & \textbf{86.82} & 73.92 & 84.78 & 62.88 & 71.84 \\
\checkmark & \checkmark & \checkmark & & 78.71 & 86.17 & 76.92 & 87.08 & 64.00 & 72.42 \\
\checkmark & \checkmark & \checkmark & \checkmark & 78.66 & 85.91  & \textbf{77.67} & \textbf{87.65} & \textbf{65.09} & \textbf{73.59} \\
\hline
\end{tabular}}
\end{table}

\begin{table}
\centering
\caption{
Our method with different cross-modal collaboration operations.
NC means node-level collaboration.
EC means edge-level collaboration.
CT means concatenation.
%
%
%
CA means cross-attention in our method.
}
\label{tab:fusion}
\resizebox{\linewidth}{!}{
\begin{tabular}{c|c|c|c|c|c|c|c}
\hline
\multirow{2}{*}{NC} & \multirow{2}{*}{EC} & \multicolumn{2}{|c|}{Object} & \multicolumn{2}{|c|}{Predicate} & \multicolumn{2}{|c}{Triplet} \\
\cline{3-8}
& & A@$1$ & A@$5$ & mA@$1$ & mA@$3$ & mA@$50$ & mA@$100$ \\
\hline
CT & CT & 55.78 & 77.58 & 51.64 & 74.13 & 60.37 & 72.66 \\
CT & CA & 56.14 & 78.38 & 52.28 & 75.04 & 61.50 & 73.80 \\
CA & CT & 56.00 & 77.68 & 52.14 & 73.54 & 63.92 & 73.10 \\
\hline
CA & CA & 55.66 & 78.66 & 54.03 & 77.67 & 65.09 & 73.59 \\
\hline
\end{tabular}}
\end{table}

\begin{table}
\centering
\caption{
Performance gains brought by our VL-SAT scheme with two reference 3DSSG prediction models.
}
\label{tab:sgg}
\resizebox{\linewidth}{!}{
\begin{tabular}{l|c|c|c|c|c|c}
\hline
& \multicolumn{2}{|c|}{Object} & \multicolumn{2}{|c|}{Predicate} & \multicolumn{2}{|c}{Triplet} \\
\cline{2-7}
& A@$1$ & A@$5$ & mA@$1$ & mA@$3$ & mA@$50$ & mA@$100$ \\
\hline
\sggpoint~\cite{zhang2021exploiting} & 51.42 & 74.56 & 27.95 & 49.98 & 45.02 & 56.03 \\
+VL-SAT & 52.08 & 75.76 & 38.04 & 60.36 & 52.51 & 64.31 \\
\hline
SGFN~\cite{wu2021scenegraphfusion} & 53.67 & 77.18 & 41.89 & 70.82 & 58.37 & 67.61 \\
+VL-SAT & 55.43 & 78.88 & 52.91 & 72.37 & 63.57 & 72.02 \\
\hline
\end{tabular}}
\end{table}

\subsection{Ablation Study and Analysis}
\label{subsec:ablation}

\noindent\textbf{Ablation Study.} 
In \cref{tab:ablation_exp}, we conduct a comprehensive ablation study.
The first row denotes the baseline method ``no-VL-SAT''.
From \cref{tab:ablation_exp}, we could observe that the
%
%
CLIP-initialized object classifier brings about $1.41\%$ gains on object A@$5$.
Node and edge-level collaboration and triplet-level CLIP-based regularization steadily bring gains on triplet prediction, with $2.23\%$, $1.12\%$, and $1.09\%$ boost on mA@$50$ metric. 
%
%
\textcolor{black}{It is worth noting that regularizing the training of predicates and triplets may bring bias to the representation of objects, which leads to a slight drop in object prediction when EC/TR is employed.
Thanks to the NC/CI modules, this degradation is not severe, which validates the effectiveness of our VL-SAT training scheme.}

\vspace{+1mm}
\noindent\textbf{Different Cross-modal Collaboration Strategies.} 
We investigate the effects of different cross-modal collaboration operations in~\cref{tab:fusion}.
We compare a simple operation named CT, in which we just concatenate corresponding features between two models.
%
%
When CT is applied, the mA@$50$ metric of the triplet prediction drops significantly.
%
%
By employing our multi-head cross-attention (termed as CA) in both collaborations, significant gains can be observed.
%


\vspace{+1mm}
\noindent\textbf{Generalization Ability.}
In \cref{tab:sgg}, we also show the performance gains brought by our VL-SAT scheme with two reference 3DSSG prediction models, namely \sggpoint~\cite{zhang2021exploiting} and SGFN~\cite{wu2021scenegraphfusion}.
VL-SAT shows consistent performance gains with different 3D prediction models, especially with respect to the evaluation of predicate and triplet.
%

\section{Conclusions}
We have introduced a visual-linguistic semantics assisted training (VL-SAT) scheme to boost 3D semantic scene graph prediction in the point cloud.
We build a strong oracle multi-modal model, which captures structural semantics using extra input data from vision, auxiliary training signals from language, and geometric features from the 3D model.
The oracle multi-modal model enhances the 3D prediction model via back-propagated gradient flows.
Consequently, the 3D prediction model can predict reliable scene graphs with only a 3D point cloud as input.
%
%
Qualitative and quantitative results demonstrate that our method remarkably outperforms the existing methods.
\label{sec:conclusion}

\vspace{+1mm}
\noindent\textbf{Acknowledgements.} This work was supported by National Key Research and Development Program of China (2021YFB1714300), and National Natural Science Foundation of China (62132001, 62233005).

{\small
\bibliographystyle{ieee_fullname}
\bibliography{egbib}
}

\end{document}


\title{VL-SAT: Visual-Linguistic Semantics Assisted Training for 3D Semantic Scene Graph Prediction in Point Cloud \\ (Supplementary Material)}

\maketitle

\setcounter{table}{0}
\setcounter{figure}{0}
\setcounter{section}{0}
\renewcommand\thesection{\Alph{section}}
\renewcommand{\thefigure}{S\arabic{figure}}
\renewcommand{\thetable}{S\arabic{table}}

\begin{figure*}[t]
\centering
\includegraphics[width=1.0\linewidth]{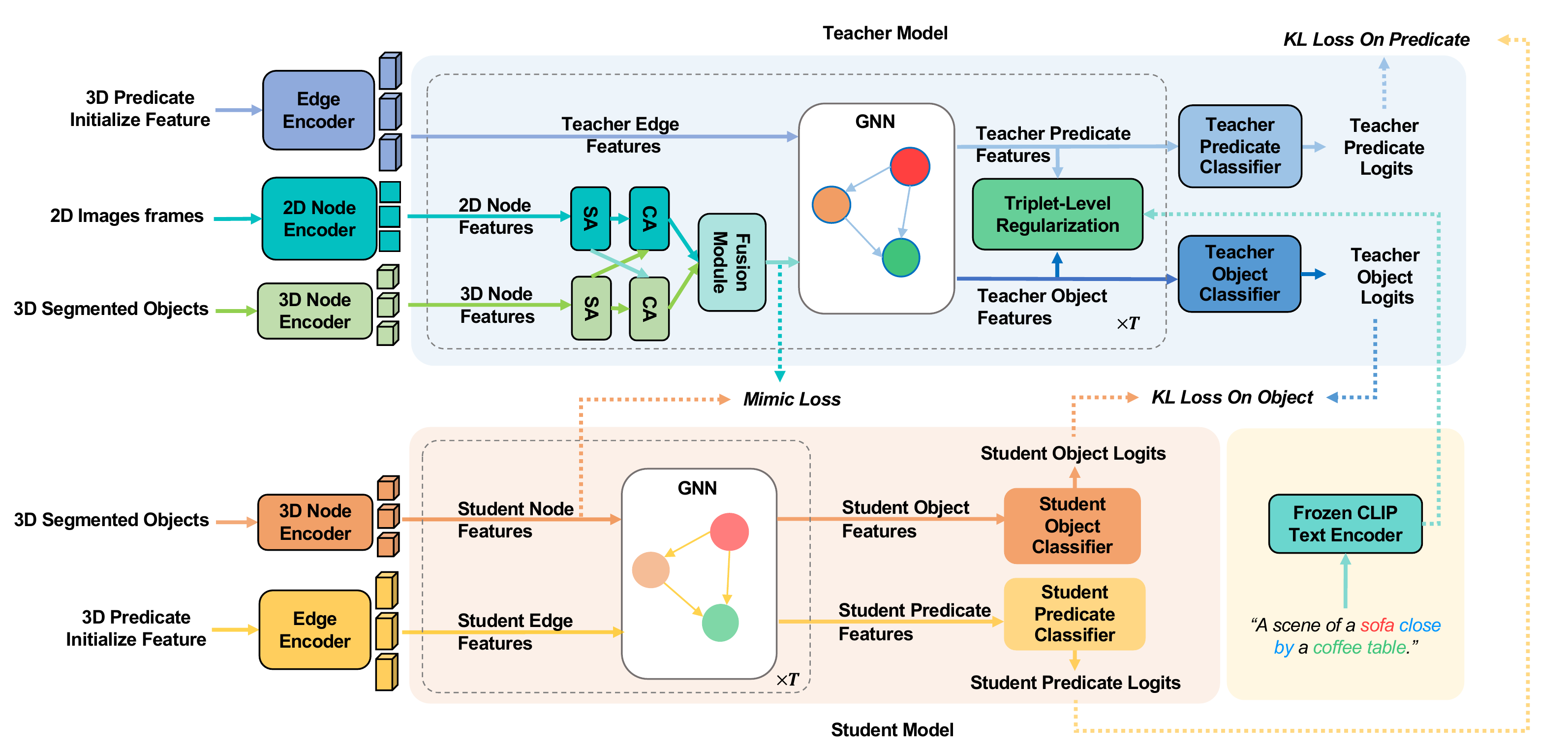}
\caption{
\textbf{The Teacher-Student Model based on the Knowledge Distillation Scheme.}
%
%
During training, the teacher model transfers its knowledge to the student model via feature mimic.
Besides, we also add KL loss between teacher logits and student logits on both object and predicate classifiers to advance the knowledge transfer process.
During inference, the student model takes the same inputs as the 3D model in our VL-SAT scheme.
}
\label{fig:network_kd}
\end{figure*}

\begin{table*}
\centering
\resizebox{\textwidth}{!}{
\begin{tabular}{c|c}
\hline
Split & Predicate \\
\hline
Head & left, right, front, behind, close by, same as, attached to, standing on \\
Body & bigger than, smaller than, higher than, lower than, lying on, hanging on \\
Tail & supported by, inside, same symmetry as, connected to, leaning against, part of, belonging to, build in, standing in, cover, lying in, hanging in \\
\hline
\end{tabular}}
\caption{Splits of predicates.}
\label{tab:splits}
\end{table*}

\begin{table}
\centering
\resizebox{\linewidth}{!}{
\begin{tabular}{c|c|c|c|c|c}
\hline 
\multirow{2}{*}{Method} & \multicolumn{3}{c|}{Predicate} & \multicolumn{2}{c}{Triplet} \\ 
\cline{2-6}
& mA@1 & mA@3 & mA@5 & mA@50 & mA@100 \\ 
\hline
SGFN                             & 41.89 & 70.82 & 81.44 & 58.37 & 67.61  \\
\hline
KD (Teacher)  & 53.57 & 72.37 & 86.18 & 73.31 & 81.08  \\
VL-SAT (Oracle)                   & 55.66 & 76.28 & 86.45 & 74.10 & 81.38  \\
\hline
KD (Student)  & 52.22 & 72.50 & 83.18 & 62.92 & 71.75  \\
VL-SAT (Ours)                     & 54.03 & 77.67 & 87.65 & 65.09 & 73.59  \\
\hline
\end{tabular}}
\caption{
Results of different knowledge transfer methods. 
We refer to the multi-modal teacher-student model as Knowledge Distillation (KD) scheme, and then we compare the results with our VL-SAT scheme.}
\label{tab:ablation_kd}
\end{table}

\begin{table}
\centering
\resizebox{\linewidth}{!}{
\begin{tabular}{c|c|c|c|c|c|c}
\hline 
\multirow{2}{*}{Model}  & \multicolumn{2}{c|}{Object}  & \multicolumn{2}{c|}{Predicate} & \multicolumn{2}{c}{Triplet} \\
\cline{2-7}
& A@5 & A@10 & mA@3 & mA@5 & mA@50 & mA@100 \\ 
\hline
\baseclip (XYZ)            & 79.03 & 86.81 & 72.50 & 83.59 & 60.65 & 69.71  \\
\baseclip (XYZ+RGB)        & 76.35 & 84.19 & 71.45 & 79.10 & 58.76 & 67.67  \\
VL-SAT(ours)            & 78.66 & 85.91 & 77.67 & 87.65 & 65.09 & 73.59   \\
\hline
\end{tabular}}
\caption{
Results of different inputs. 
%
%
We figure out whether adding RGB information directly into the 3D point cloud (XYZ) input can boost 3DSSG prediction performance as our VL-SAT scheme does.
\baseclip shares the same network architecture as non-VL-SAT but leverages CLIP-initialized object classifier.
}
\label{tab:ablation_input}
\end{table}

\begin{table}
\centering
\resizebox{\linewidth}{!}{
\begin{tabular}{c|c|c|c|c|c}
\hline 
\multirow{2}{*}{Backbone} & \multicolumn{3}{c|}{Predicate} & \multicolumn{2}{c}{Triplet} \\ 
\cline{2-6}
& mA@1 & mA@3 & mA@5 & mA@50 & mA@100 \\ 
\hline
non-VL-SAT & 41.99 & 70.88	& 81.67 & 59.58 & 67.75  \\
CLIP Pretrained & 48.65 & 76.12 & 87.09 & 62.85 & 71.60  \\
ImageNet21k Pretrained & 47.43 & 74.47 & 85.71 & 61.36 & 70.07 \\
\hline
\end{tabular}}
\caption{
Results of different visual encoders. 
%
%
We figure out the influence of visual assistance and the influence of visual encoder pretrained using different datasets.
We conduct the experiments with a variant of the VL-SAT scheme, which discards all the linguistic assistance, \ie CLIP-based object classifier initialization, and CLIP-based triplet-level regularization.
}
\label{tab:ablation_encoder}
\end{table}


\section{Implementation Details}
\vspace{+1mm}
\noindent\textbf{2D Data Preparation.}
Since each 3D scan in the 3DSSG dataset~\cite{wald2020learning} is associated with RGB sequences with known camera poses, thus it is possible to extract 2D image patches associated with each point cloud instance $\mP_{i}$.
We first project the 3D points in $\mP_{i}$ to each RGB frame according to the given camera pose, and then calculate the area of the enlarged bounding box surrounded by the projected points. Since then, we rank the frames in the descending order of these areas and select the image patches in the bounding boxes in the top-$N$ frames as the N-view image patches of the instance $\mP_i$.
The visual features $\vo_i$ corresponding to $\mP_i$ are thus generated by mean pooling the visual features of N-view image patches through a fixed CLIP vision encoder that has been finetuned on 3DSSG~\cite{wald2020learning,gao2021clip}.

%
%
%
%

\vspace{+1mm}
\noindent\textbf{Architecture Details.}
We adopt a simple PointNet~\cite{qi2017pointnet} as the 3D node encoder.
%
%
%
As for the 2D node encoder, we use Vit-B-32 architecture~\cite{dosovitskiy2021an} as the backbone of the CLIP image encoder.
%
%
The feature dimension of all the node and edge features in the oracle and 3D model is set to be $512$.
%
%
The structure of GNN is borrowed from SGFN\cite{wu2021scenegraphfusion}, which uses a FAT mechanism to combine neighboring features.
All the multi-head self attention (MHSA) or multi-head cross attention (MHCA) structures in our method use $8$ heads, with a hidden feature size of $512$.
According to our experiments, $\rho(\cdot,\cdot)$ in $L_\text{tri-emb}$ is implemented with $\ell_1$ norm, and the $\rho(\cdot,\cdot)$ in $L_\text{node-init}$ is implemented with negative cosine distance.

\vspace{+1mm}
\noindent\textbf{Splits of Predicates.} 
We split the $26$ predicate classes into three parts: \textit{head}, \textit{body}, \textit{tail}.
In detail, we sort the predicates according to their frequencies in the training set in descending order and select the top $8$ categories as head classes, the last $12$ categories as tail classes, and the remaining $6$ categories as body classes.
You can refer to~\cref{tab:splits}.

\section{More Experiments}
\subsection{Comparison with Knowledge Distillation Scheme.}
To prove the superiority of our proposed VL-SAT scheme, we design a knowledge distillation (KD) scheme as in ~\cref{fig:network_kd}, which adheres to a teacher-student paradigm.
The teacher is a multi-modal model, which fuses visual and geometrical information using bi-directional cross-attention.
Besides, to compare with our VL-SAT scheme in a fair manner, we also leverage linguistic assistance in the KD scheme.
The student model is the same as our non-VL-SAT model.
The knowledge transfer process from teacher to student is implemented with traditional mimic loss and KL loss.
As shown in ~\cref{tab:ablation_kd}, since our oracle model trained with VL-SAT scheme can combine multi-modal knowledge more effectively, the performance is better than the teacher model of KD scheme among all metrics, \eg $2.1\%$ gains on predicate mA@$1$.
%
%
Besides, VL-SAT (ours) outperforms KD (student) with $2.1\%$ gains on triplet mA@$50$.
We think the performance degradation of KD scheme is because
%
%
the teacher model has a different network structure compared with the student model, and the heterogeneous network structures may hinder the knowledge transfer process as indicated in ~\cite{wang2022head}.

\subsection{Can RGB Information on Point Cloud Boost 3DSSG Prediction As Well?}
%
%
%
%
%
%
%
%
Since the VL-SAT scheme boosts 3DSSG prediction significantly, it is intuitive to think about whether adding RGB information directly into 3D point cloud could also do well.
We conduct such experiments (namely, $\text{Base}_\text{CLIP}$ since we employ ClIP-initialized object classifier in this baseline) in \cref{tab:ablation_input} and find that simply concatenating RGB values to point cloud's XYZ coordinates (as $\text{Base}_\text{CLIP}$ (XYZ+RGB)) brings moderate performance drop (as $\text{Base}_\text{CLIP}$ (XYZ)) in 3DSSG prediction task.
We doubt it is due to over-fitting on RGB values as indicated in ~\cite{qi2020imvotenet}.
The experiment results also validate the necessity of our VL-SAT scheme.

\subsection{Influence of Visual Assistance.}
%
%
%
%
%
%
%
%

To investigate the influence of visual assistance, we conduct experiments without linguistic assistance, \ie CLIP-based object classifier initialization, CLIP-based triplet-level regularization, during training.
As shown in ~\cref{tab:ablation_encoder}, with only visual assistance, our method still obtains $6.66\%$ gain on predicate mA@$1$ and $3.27\%$ gain on triplet mA@$50$.
Furthermore, we try the visual encoder pretrained on ImageNet21K~\cite{deng2009imagenet} dataset, which shares the same network structure as the CLIP pretrained visual encoder used in our VL-SAT.
The ImageNet21K pretrained visual encoder also shows performance gains over non-VL-SAT model, but is inferior to our CLIP pretrained visual encoder.
The result shows that the CLIP pretrained visual encoder possesses a stronger representation ability over the ImageNet21K pretrain visual encoder.

%
%
%
%


{\small
\bibliographystyle{ieee_fullname}
\bibliography{egbib}
}